\documentclass[a4paper, 10pt, conference]{ieeeconf}      

\usepackage{times}
\usepackage{epsfig}
\usepackage{graphicx}
\usepackage{amsmath}
\usepackage{amssymb}
\usepackage{textcomp}
\usepackage{multirow}
\usepackage{adjustbox}

\usepackage{multirow}
\usepackage{colortbl}
\usepackage{hhline}
\usepackage{subcaption}
\usepackage{verbatim}

\usepackage{gensymb}
\usepackage{flushend}
\usepackage{xcolor}
\usepackage{hyperref}

\IEEEoverridecommandlockouts                              

\title{\LARGE \bf
SoilingNet: Soiling Detection on Automotive Surround-View Cameras 
}

\author{Michal U\v{r}i\v{c}\'{a}\v{r}$^{1}$, Pavel K\v{r}\'{i}\v{z}ek$^{1}$, Ganesh Sistu$^{2}$ and Senthil Yogamani$^{2}$ \\ 
$^{1}$Valeo R\&D Prague, Czech Republic $^{2}$Valeo Visions Systems, Ireland \\
         {\tt \small \{michal.uricar, pavel.krizek, ganesh.sistu, senthil.yogamani\}@valeo.com}
}

\begin{document}


\maketitle
\thispagestyle{empty}
\pagestyle{empty}

\begin{abstract}
Cameras are an essential part of sensor suite in autonomous driving. Surround-view cameras are directly exposed to external environment and are vulnerable to get soiled. Cameras have a much higher degradation in performance due to soiling compared to other sensors. Thus it is critical to accurately detect soiling on the cameras, particularly for higher levels of autonomous driving. We created a new dataset having multiple types of soiling namely opaque and transparent. It will be released publicly as part of our WoodScape dataset \cite{yogamani2019woodscape} to encourage further research.
We demonstrate high accuracy using a Convolutional Neural Network (CNN) based architecture. We also show that it can be combined with the existing object detection task in a multi-task learning framework. Finally, we make use of Generative Adversarial Networks (GANs) to generate more images for data augmentation and show that it works successfully similar to the style transfer.
\end{abstract}

\section{INTRODUCTION}

Autonomous driving systems are becoming mature by using a variety of different sensors. Cameras continue to be one of the key sensors as the road infrastructure is designed for human visual sensors. The first generation systems primarily used a single camera and more recently more cameras are used to get full coverage around the vehicle to handle more complex driving scenarios \cite{heimberger2017computer}. 
Horgan et al. \cite{horgan2015vision} provides an overview of various visual perception tasks prior to deep learning era and Sistu et al. \cite{sistu2019neurall} provide an overview from a deep learning perspective. \\

Quality of computer vision algorithms degrade significantly in poor environmental conditions due to bad weather including rain, fog, snow, etc. This is particularly worsened when the camera lens is exposed to rain or it becomes frozen in winter. In addition, external cameras are exposed to mud and dust as well. Thus it is important to detect when camera lens is soiled so that the system is aware the vision algorithms will degrade severely. In previous generation systems with lesser automation like Level 2 \cite{SAE_automation}, soiling detection was done to let the human driver know that vision algorithms are less reliable. For higher levels of automation like Level 4, the self driving system needs to automatically detect soiling and also correct for it, e.g. by using a cleaning system. \\ 

\begin{figure*}[tb]
    \centering
    \includegraphics[width=0.89\linewidth]{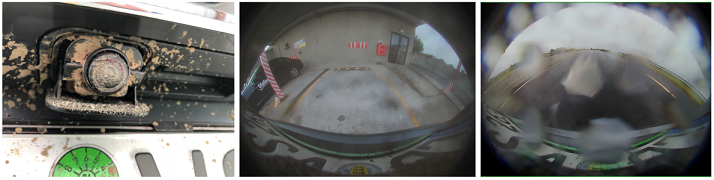}
    \caption{From left to right: a) soiled camera lens mounted to the car body; b) the image quality of the soiled camera from the previous image; c) an example of image soiled by a heavy rain.}
    \label{fig:saw_definition}
\end{figure*}

Soiling of camera lens occurs in mobile phones as well and modern mobile phones detect them and ask the user to clean it manually. However, there is very little literature and datasets available for soiling detection. The closest problem to the soiling detection addressed in literature is in recent work of Porav et al. \cite{porav2019i}, where they created a setup to capture water dripping on camera lens and have a parallel setup without water drops. However, in our experience for soiling by water, rain drops splashes hard on the lens as illustrated in Fig. \ref{fig:saw_definition} (c) and slow dripping does not capture it. A related problem is the design of robust algorithms to implicitly handle these scenarios, for example Sakaridis et al. \cite{sakaridis2018semantic} developed a robust semantic segmentation algorithm to handle foggy scenes. An alternate approach is to image restoration to improve the quality of the image, for example Dehazing approaches were proposed in \cite{Fattal-2008}, \cite{Berman-2016} and \cite{Ki-2018}. More recently, the computer vision conference CVPR 2019 has organized a competition called UG2 \cite{ug2} to evaluate vision algorithms in poor visibility conditions for autonomous driving. \\

\noindent The main contributions of this paper are as follows: 
\begin{enumerate} 
    \item Introduction and formal definition of soiling detection task in automotive scenarios.
    \item Implementation of a high accuracy soiling detection algorithm and experimental results.
    \item Incorporation of Generative Adversarial Networks (GAN)~\cite{Goodfellow-NIPS2014} based data augmentation to improve results.
    \item Release of the first public soiling dataset briefly introduced in our WoodScape dataset paper \cite {yogamani2019woodscape}.
\end{enumerate} 


The paper is organized as follows. Section \ref{sec:soiling} defines soiling detection task formally, discusses an automated cleaning system and design of a soiling dataset.
Section \ref{sec:proposed} discusses the experimental setup, proposed Convolutional Neural Network (CNN) architecture and GAN based data augmentation. Section \ref{sec:results} presents the experimental results and analysis. Finally, Section \ref{sec:conc} summarizes and concludes the paper.

\section{SOILING DETECTION TASK} \label{sec:soiling}

\subsection{Motivation}

As discussed in the introduction, soiled lens degrades vision algorithm and it is critical to notify the system and/or the  driver about potential degradation impact. More specificaly, soiling detection output is used for three aspects of autonomous driving system discussed below.\\

\begin{figure}[htpb]
    \centering
    \includegraphics[width=0.95\linewidth]{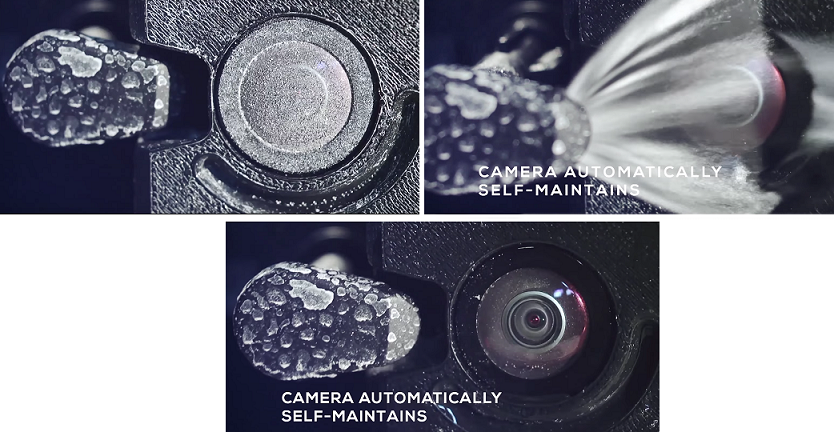}
    \caption{Illustration of our camera cleaning system}
    \label{fig:cleaning}
\end{figure}

\textbf{Cleaning System:} Surround view cameras are usually directly exposed to the adverse environmental conditions. Therefore, one cannot avoid a situation when e.g. a splash of mud or other kind of dirt hits the camera. Another, even more common example would be a heavy rain when the water drops frequently hit the camera lens surface. As the functionality of visual perception degrades significantly, detection of soiled cameras is necessary for achieving higher levels of automated driving~\cite{SAE_automation}.  Fig. \ref{fig:cleaning} illustrates our cleaning system which clears ice by spraying warm water. Some systems additionally have an air blower to clean up remaining water. As the cleaning system is water based, its water tank needs to be refilled periodically. Thus, it is important to have an algorithm with low false positives to reduce the refilling. \\

\textbf{Algorithm Degradation:}
Soiling detection can be used for estimation of algorithm degradation to disable partially or reduce confidence of the output.  The simple degradation mechanism is to switch off all the algorithms if there is any soiling on the lens. However in practice, it is important to support partial degradation of the algorithms. For example, the soiling detection algorithm may output severity of soiling which can be used to reduce the confidence of the visual perception algorithms so that the outputs are still useful. It is also useful to localize the soiling instead of image level soiling detection because certain parts of the image may still be clean and perception algorithms may work fine. Many classical computer vision algorithms are local operators and thus can be disabled selectively for certain soiled regions. However, CNNs are global operators and their behavior to partial soiling is not studied in literature. \\

\textbf{Image Restoration:}
Image restoration is employed in autonomous driving to improve the quality for conditions such as motion blurring, low light and other environmental degradation such as rain, haze, etc. Some of the restoration techniques are de-raining \cite{porav2019i}, de-fogging \cite{sakaridis2018semantic} and de-hazing \cite{Ki-2018}. Specifically in case of soiled lens, the restoration techniques can be specialized for the particular soiling type and thus soiling detection can be useful. For example, when transparent soiling is detected, specialized restoration technique for inpainting these regions can be used by leveraging partially visibility.

\subsection{Formal definition}


To our best knowledge, the soiling detection task was so far only briefly described in our previous paper~\cite{Uricar-2019a}. However, in that paper, we only outlined the problem and described the possibilities of applying GAN~\cite{Goodfellow-NIPS2014} for an advanced data augmentation related to the soiling imagery. In this paper, we would like to define the problem of soiling detection/classification and to present a solution and report the empirical results. 

We define the camera soiling detection task as a multi-label classification problem. Each image can be described by a binary indicator array, where zeros, and ones correspond to the absence or presence of a specific soiling class, respectively. The soiling classes we define are $\mathcal{C} = \{ \mathrm{opaque}, \mathrm{transparent} \}$. And the definition of each class is as follows: as an ``opaque'' class we label such regions in the image, where it is impossible to say what would be seen if there was no soiling, i.e. the regions which are coloured by nontransparent colors preventing to see what is behind; as a ``transparent'' class we label regions which are apparently blurred or deformed from the expected appearance, however it is possible to distinguish colors from the original scenery, i.e. it is possible to see ``behind''. Note, that such definition makes some seemingly unnatural structures possible, such as an ``opaque'' water soiling. However, from the data which we have recorded for this purpose, we can say that such structures are quite frequently observed, which justifies our definition.



Given a single image $I \in \mathcal{I}$, we are interested in a classifier $g: \mathcal{I} \rightarrow \mathcal{C}^2$, where $\mathcal{C}$ denotes the set of class labels: $\mathcal{C} = \{ \mathrm{opaque}, \mathrm{transparent} \}$, where the labels are supposed to be binary, specifying if the given type of soiling (i.e. opaque or transparent) is present (in such case the value is $1$) or not (the value is $0$). Then clearly a vector $\mathbf{c_1} = [0, 0]$ denotes a clean image, while $\mathbf{c_2} = [1, 1]$ denotes an image with both type of soiling categories present. The sought classifier $g$ should be optimal with respect to minimization of the error, which is measured by the number of mis-classifications of the soiling categories. We measure the mis-classifications by Hamming distance of the binary indicator arrays of the ground truth manually annotated labels and the predictions returned by the classifier:
\begin{equation}
    \varepsilon_{\mathrm{cat}}(\mathbf{c_{\mathrm{GT}}}, \mathbf{c}) = \mathrm{ham}(\mathbf{c_{\mathrm{GT}}}, \mathbf{c})
\end{equation}
%
where $\mathbf{c}$ denotes the predicted binary encoded category, $\mathbf{c_{\mathrm{GT}}}$ is the corresponding ground truth label. Finally,  $\mathrm{ham}(\cdot, \cdot)$ denotes the Hamming distance. The overall error is measured as the average value of $\varepsilon_{\mathrm{cat}}(\mathbf{c_{\mathrm{GT}}}, \mathbf{c})$ over the whole testing set.

\begin{figure}[t]
    \centering
    \includegraphics[width=0.5\textwidth]{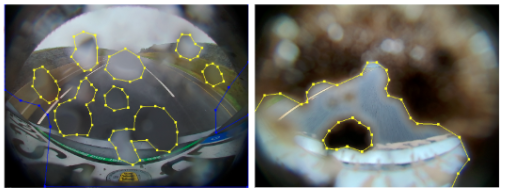}
    \includegraphics[width=0.4\textwidth ,trim={0 2cm 0 0},clip]{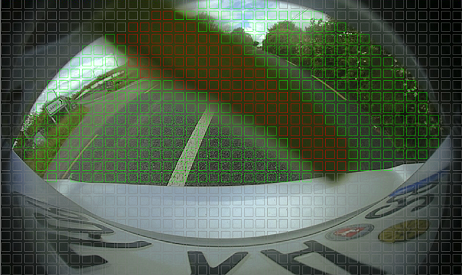}
    
    \caption{Soiling annotation using polygons (top) and tile level ground truth generation derived from polygons (bottom) }
    \label{fig:soilingannotation}
\end{figure}

The above definition of image level classification can be generalized to tile-level for providing higher level of spatial resolution of detection as illustrated in Fig. \ref{fig:soilingannotation} (bottom image).  When the tile size is 1x1, it specializes to pixel-level labelling semantic segmentation task. When the tile size is equal to image height by image width, it becomes image classification task. In a classical feature extraction plus classifier setting, each tile can be independently processed but with deep learning models global context is leveraged for output of each tile

\section{DATASET DESIGN AND PROPOSED ARCHITECTURE} \label{sec:proposed}

This section summarizes the main contributions of the paper in the following three sub-sections. Firstly we created a dataset which will be partially shared publicly as it is a newly defined task in autonomous driving. We demonstrate successful application of GAN based data augmentation technique. Finally, we discuss supervised CNN models for soiling detection and demonstrate good accuracy.

\begin{figure*}
    \centering
    \includegraphics[width=0.75\textwidth]{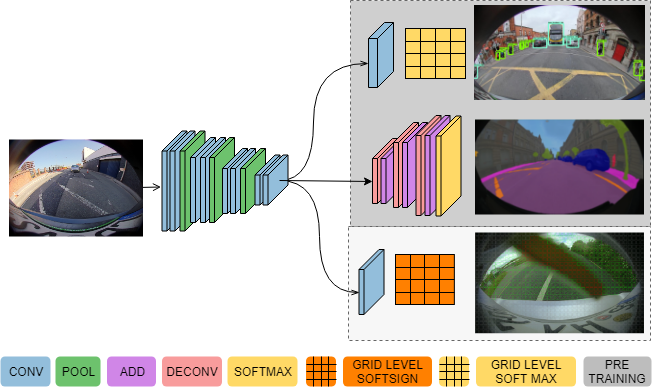}
    \caption{Illustration of soiling integrated into multi-task object detection and segmentation network.}
    \label{fig:multi-stream-task}
\end{figure*}

\subsection{Dataset Creation}


Altogether, we have collected $76,448$ number of images, our sampling strategy was simply to extract each $10$-th frame out of short video recordings. This is done mainly in order to decrease the self similarity of images in the dataset. 
The annotations were created manually, by clicking points of a  coarse polygonal segmentation of the soiling contained in the image with an additional information about the qualities of the soiling (if it is transparent or opaque). Note, that such type of annotations allows us to use the same representation for both whole-frame based and tile-based detection experiments, since the polygonal annotation can be converted to the requested annotation vector format straightforwardly, i.e. the polygonal annotation (where each plygon is marked to be either fully transparent or fully opaque) is converted to the tile-based label\footnote{Recall, that the definition of tiles size might boil down to a singular tile covering the whole image.}, where the presence of each soiling class is calculated based on the percentage cover of the tile overlap with the corresponding polygon.

For the need of training, model selection and final experimental evaluation of the trained classifiers, we split the dataset to non-overlapping training, validation, and test sets. We used the golden standard of $60$/$20$/$20$ ratio for obtaining these splits and also we used stratified sampling approach, to retain the underlying distributions of the classes among the splits. This process was done independently for the whole-frame based approach and for the tile-based approach, respectively. 

This leads to the following statistics. We have collected $45,868$ and $15,291$ number of images for training and evaluation purpose, respectively. The training dataset annotation statistics are as follows: $22,015$ frames are clean (i.e. their annotation is a vector $[0, 0]$), $11,704$ are containing opaque soiling only (annotation is $[1, 0]$), $4,623$ contain transparent soiling only ($[0, 1]$), and, finally, $7,526$ are containing both kind of soiling classes ($[1, 1]$). The testing dataset annotation statistics are as follows: $7,339$ frames are clean, $3,902$ are containing opaque soiling only, $1,541$ contain transparent soiling only, and finally, $2,509$ contain both kind of soiling classes.
We plan to provide a subset of $5,000$ images to the community, to encourage the further research on this topic.

\subsection{GAN based data-augmentation}


Getting relevant data for the soiling classification task is a very tedious task. First problem are the suitable conditions for increasing the probability that the soiling event might even occur. This makes the problem hard, but still solvable, one just needs to find a way how to manually ``soil'' the camera lens, while retaining the characteristic appearance of the realistic scenes. Another problem is the annotation of such imagery, which is extremely expensive and time consuming. 

Due to the above mentioned reasons, we used GAN to alleviate the lack of relevant data. The idea is simply that we would like to use GAN machinery for creating soiled images from the clean images and use this concept for data augmentation. Additionally, usage of GAN has the potential of semi-supervised and unsupervised learning, which we want to pursue in the future work.

Because of the problem of expensive annotations, we first tried out the CycleGAN~\cite{Zhu-ICCV-2017} approach. Since CycleGAN requires just a simple categorization of data into $2$ categories. The images can be totally unaligned (thanks to the cycle-consistency loss involved in the optimization criterion), which is yet another benefit as one can pre-process his data very easily. In our case these categories are ``clean'', and ``soiled'' images, respectively. Therefore, the CycleGAN's output of our experiments are $2$ generators, one trying to introduce the soiling into the image and the other trying to remove it and making the image ``clean''.

The main problem of CycleGAN experiment is the inability to produce variable output. Therefore, we have pracitcally no control over the generated image. This lead us to conducting another experiment with the MUNIT~\cite{huang2018multimodal} approach. The Multi-modal UNsupervised Image to image Translation (MUNIT) separates the content in the image from its style and via this we have the possibility to control the image appearance by changing the style provided at the image generation step.

\subsection{Proposed CNN architecture} 
Our goal is to design an efficient soiling detection architecture which can be deployed on a low power embedded platform having computational processing power of 1 Tera OPS. Soiling detection is an additional module in the autonomous driving platform which already contains CNN based object detection and segmentation and other classical computer vision modules like depth estimation. Thus we propose to leverage existing CNN features in the system by sharing the encoder for a soiling decoder in a multi-task network illustrated in Fig. \ref{fig:multi-stream-task}. We show that this will provide higher efficiency in the overall system relative to having a separate network for soiling.
The details of the baseline multi-task network are shared in our previous paper \cite{sistu2019real}. It comprises of a simplified ResNet10 like encoder and two decoders namely simplified YOLO V2 for object detection and simplified FCN8 for segmentation.

In this work, we propose to add the new soiling decoder as a third task. The output of the soiling decoder is tiled soiling class output. The number of tiles is set at training time. We made use of two configurations, one in which the tile size was $64\times64$ and in the other case the output was at image resolution. The three losses were scalarized into one loss using a weighted average and the weights were optimized by hyper-parameter tuning using grid search. 
As an ablation study to evaluate soiling detector on its own without multi-task network, we removed the other decoders and evaluated the network for soiling task. Soiling decoder has two convolution layers and a final grid level softsign layer which produces prediction of soiling type for each grid. In case of image level soiling experiments, decoder has the same convolution layers but with a higher stride to reduce the spatial dimensions less gradually. We use binary cross entropy per class as the loss function. The implementation was done using Keras \cite{chollet2015keras}. ADAM optimizer was used as it provides faster convergence with a learning rate of 0.0005. The input image resolution is 1280x800.

\section{EXPERIMENTAL RESULTS} \label{sec:results}

\begin{table}[tb]
\centering
\resizebox{0.3\textwidth}{!}{
\begin{tabular}{|l|c|c|}
\hline
\multicolumn{1}{|c|}{\textbf{Camera}} & \textbf{Precision} & \textbf{Recall} \\ \hline
Front                             & 59.41\%         & 99.16\%            \\ \hline
Rear                              & 64.17\%         & 99.94\%            \\ \hline
Right                          & 71.03\%         & 98.89\%            \\ \hline
Left                           & 69.99\%         & 99.72\%            \\ \hline
\end{tabular}
}
\caption{Tile-level soiling classification accuracy}
\label{tab:tiles}
\end{table}

\begin{table}[tb]
\centering
\resizebox{0.45\textwidth}{!}{
\begin{tabular}{|l|c|c|}
\hline
\multicolumn{1}{|c|}{\textbf{Encoder Training Strategy}} & \textbf{TPR} & \textbf{FPR} \\ \hline
Single-task - ImageNet pre-training                                    & 52\%     &   16\%           \\ \hline
Single-task - No pre-training                                    & 56\%            &  14\%            \\ \hline
Multi-task training                                    &  55\%             &   14\%            \\ \hline
Mutlti-task with GAN augmentation                                         & 58\%            & 14\%           \\ \hline
\end{tabular}
}
\caption{Evaluation of different training strategies for image-level soiling classification (metrics rounded off)}
\label{tab:training}
\end{table}

\begin{table*}[tb]
\centering
\resizebox{0.8\textwidth}{!}{
\begin{tabular}{|l||c|c|c|c||c|c|c|c|}
\hline
                       & \multicolumn{4}{c||}{Normalized Confusion Matrix}                                                                        & \multicolumn{4}{c|}{Raw Confusion Matrix}                                                                               \\ \hline
\multicolumn{9}{|c|}{\textbf{Training and Testing on front/rear cameras}}                                                                                                                                                                                                               \\ \hline
\multicolumn{1}{|c||}{--} & \multicolumn{1}{l|}{Clean} & \multicolumn{1}{l|}{Transparent} & \multicolumn{1}{l|}{Opaque} & \multicolumn{1}{l||}{Both} & \multicolumn{1}{l|}{Clean} & \multicolumn{1}{l|}{Transparent} & \multicolumn{1}{l|}{Opaque} & \multicolumn{1}{l|}{Both} \\ \hline
Clean                  & 1                          & 0                                & 0                           & 0                         & 14600                      & 0                                & 17                          & 0                         \\ \hline
Transparent            & 0.06                       & 0.92                             & 0                           & 0.02                      & 225                        & 3730                             & 13                          & 91                        \\ \hline
Opaque                 & 0.02                       & 0.02                             & 0.94                        & 0.03                      & 135                        & 134                              & 7913                        & 240                       \\ \hline
Both                   & 0                          & 0                                & 0.04                        & 0.94                      & 0                          & 9                                & 191                         & 4818                      \\ \hline
\multicolumn{9}{|c|}{\textbf{Training on front/rear cameras and Testing on all cameras}}                                                                                                                                                                                                \\ \hline
\multicolumn{1}{|c||}{--} & \multicolumn{1}{l|}{Clean} & \multicolumn{1}{l|}{Transparent} & \multicolumn{1}{l|}{Opaque} & \multicolumn{1}{l||}{Both} & \multicolumn{1}{l|}{Clean} & \multicolumn{1}{l|}{Transparent} & \multicolumn{1}{l|}{Opaque} & \multicolumn{1}{l|}{Both} \\ \hline
Clean                  & 0.7 & 0 & 0.28 & 0.02                         & 20563 & 1 & 8280 & 510                \\ \hline
Transparent            & 0.14 & 0.52 & 0.29 & 0.06                      & 1145 & 4192 & 2337 & 463                  \\ \hline
Opaque                 & 0.02 & 0.01 & 0.92 & 0.06                      & 255 & 209 & 15401 & 952               \\ \hline
Both                   & 0.01 & 0 & 0.12 & 0.87               & 73 & 14 & 1206 & 8441                    \\ \hline
\multicolumn{9}{|c|}{\textbf{Training on all cameras and Testing on all cameras}}                                                                                                                                                                                                \\ \hline
\multicolumn{1}{|c||}{--} & \multicolumn{1}{l|}{Clean} & \multicolumn{1}{l|}{Transparent} & \multicolumn{1}{l|}{Opaque} & \multicolumn{1}{l||}{Both} & \multicolumn{1}{l|}{Clean} & \multicolumn{1}{l|}{Transparent} & \multicolumn{1}{l|}{Opaque} & \multicolumn{1}{l|}{Both} \\ \hline
Clean                  & 1                          & 0                                & 0                           & 0                         & 29333 & 11 & 8 & 2              \\ \hline
Transparent            & 0 & 0.99 & 0 & 0            & 25 & 8073 & 1 & 38            \\ \hline
Opaque                 & 0 & 0 & 1 & 0             & 0 & 0 & 16816 & 1                 \\ \hline
Both                   & 0 & 0 & 0 & 1             & 0 & 13 & 14 & 9707               \\ \hline
\end{tabular}
}
\caption{Summary of results of Image-level soiling classification}
\label{tab:results}
\end{table*}

In this section, we discuss our experimental results summarized in three tables. Table \ref{tab:results} summarizes three sets of experiments done on image level classification. Table \ref{tab:tiles} summarizes tile level experiments on the four cameras individually. Table \ref{tab:training} lists the difference in accuracy by using various encoder training strategies. We discuss the results in more detail below. \\

\textbf{Image Level Classification:} The number of annotation samples is much higher at image level and we tabulate more detailed results for these experiments. The perspective of right and left cameras are very different from front and rear cameras. 
The latter is closer to the typical front camera perspective of public datasets. Thus we perform ablation studies with and without right/left cameras summarized in Table \ref{tab:results}. Both normalized and un-normalized confusion matrices are shared. There are three sets of experiments - (1) training and testing only on front and rear cameras, (2) training on front/rear cameras and testing on all cameras and (3) training on all cameras and testing on all cameras. In experiment (1), high detection rate was obtained for each class. Experiment (2) shows degradation in accuracy indicating that the network doesn't generalize from front/rear to left/right network. Experiment (3) shows that when the same network is trained on all images, it performs well and slightly better than (1) as there is better regularization.\\

\textbf{Tile Level Classification:} We perform tile-level classification using the configuration of $64 \times 64$. Precision and recall values are summarized in Table \ref{tab:tiles}. Unlike high accuracy values of image level task, precision values obtained here show that there is a lot of room for further improvement in accuracy. Front and rear have a higher probability of getting soiling especially with difficult scenarios and thus it has much higher samples. 


\textbf{Encoder Training Strategies:} We evaluate different encoder training strategies which are summarized in Table \ref{tab:training}. As discussed before, the goal is finally to design an efficient network which can be deployed on a commercial platform and we propose to make use of multi-task learning. As an ablation study, we compare it with single-task soiling only learning and ImageNet only pre-trained network. ImageNet pre-trained encoder is worse than soiling single-task encoder by 4\% in True Positive Rate (TPR) and 2\% in False Positive Rate (FPR). Multi-task encoder achieves close to single-task network encoder with a 1\% reduction in TPR, however this comes at a much higher system level efficiency. Multi-task encoder is further improved by 3\% by making using of additional GAN generated images.
\\

\begin{figure*}[htpb]
    \centering
    \includegraphics[width=0.99\linewidth]{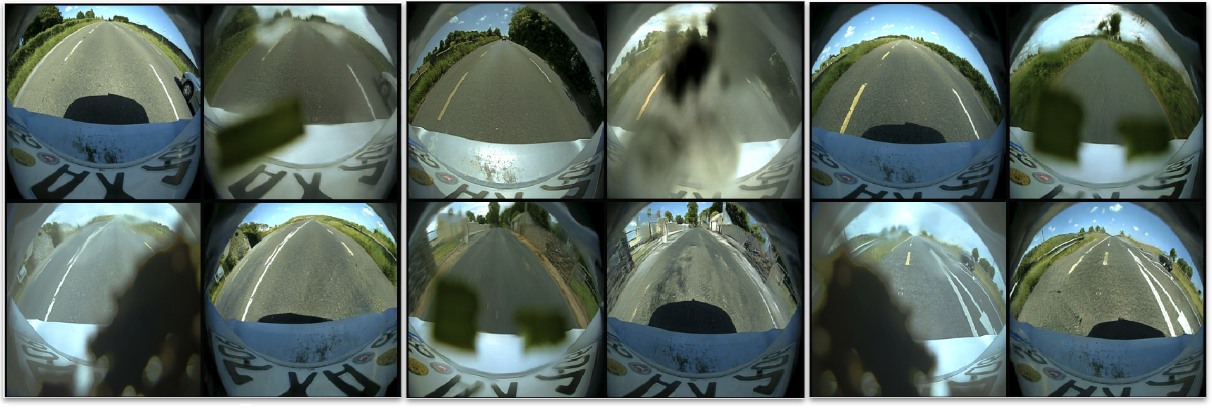}
    \caption{The results obtained from the CycleGAN proof of concept for the problem of soiling and adverse weather classification. 
    The legend per each quadruple of images: the top left image shows the original clean image; the top right image is the generated ''soiled`` version of this clean image. The bottom left image is the original soiled image; the bottom right image is the ''clean`` version of this soiled image. 
    }
    \label{fig:cycleGAN_results}
\end{figure*}

\textbf{GAN based data augmentation:}
Just after a few epochs the CycleGAN experiment started to show promising results,  it was visible that the generators are able to recognize correctly which parts of the image are affected by soiling. After several tens of epochs, we already had generator which was capable of introducing soiling into the image. The ``de-soiling'' generator needed much more time to learn. However, close to the final number of epochs, which we set to $200$, it was also doing quite good job. Note, that the ``de-soiling'' task is much harder, as the network has to learn to in-paint some characteristic structures which are fitting into the original image seamlessly. In Fig. \ref{fig:cycleGAN_results} we depict several examples of the CycleGAN experiments results.

Unfortunately, we were not able to converge with MUNIT experiments to something reasonable as it requires more parameter tuning and probably also higher quality data. The generated images were containing still too much artifacts and in some cases failing completely to generate reasonable images. 
We want to pursue with MUNIT experiments in our future work as we believe that it is a promising approach to be explored further. \\


\textbf{Practical Challenges:}
Soiling detection poses many challenges in practice. It might appear to be a standard segmentation problem however there is no spatial or geometric structure present in soiling pattern unlike other objects. Even the texture can vary drastically across different occurrence of soiling. Thus it makes it difficult for learning discriminative features. In particular, transparent soiling is challenging to detect as it subtly changes the appearance of the scene. Motion blurred areas which commonly occur in high speed scenes often becomes classified as transparent soiling. The class annotation also has to be appropriately labelled according to severity of the soiling, this can be highly subjective. 

\section{CONCLUSIONS} \label{sec:conc}

In this paper, we discussed the importance of soiling detection task and why its necessary for achieving higher levels of autonomous driving. We created a multi-camera surround view soiling dataset and a portion of this dataset will be made public to encourage further research in this area. We demonstrated that a relatively small-sized CNN achieves accurate results. We also demonstrated that GAN based augmentation using CycleGAN and MUNIT works well for soiling generation. We evaluated various training methodologies and demonstrated that soiling can be merged into an object detection and segmentation multi-task network for high efficiency.






\bibliographystyle{ieee}
\bibliography{references/egbib}

\end{document}